# Policy Iteration for Factored MDPs

**Daphne Koller**
Computer Science Department
Stanford University
Stanford, CA 94305-9010
*koller@cs.stanford.edu*

**Ronald Parr**
Computer Science Department
Stanford University
Stanford, CA 94305-9010
*parr@cs.stanford.edu*

## Abstract

Many large MDPs can be represented compactly using a dynamic Bayesian network. Although the structure of the value function does not retain the structure of the process, recent work has suggested that value functions in factored MDPs can often be approximated well using a *factored value function*: a linear combination of *restricted* basis functions, each of which refers only to a small subset of variables. An approximate factored value function for a particular policy can be computed using approximate dynamic programming, but this approach (and others) can only produce an approximation relative to a distance metric which is weighted by the stationary distribution of the current policy. This type of weighted projection is ill-suited to policy improvement. We present a new approach to value determination, that uses a simple closed-form computation to compute a least-squares decomposed approximation to the value function *for any weights* directly. We then use this value determination algorithm as a subroutine in a policy iteration process. We show that, under reasonable restrictions, the policies induced by a factored value function can be compactly represented as a decision list, and can be manipulated efficiently in a policy iteration process. We also present a method for computing error bounds for decomposed value functions using a variable-elimination algorithm for function optimization. The complexity of all of our algorithms depends on the factorization of the system dynamics and of the approximate value function.

## 1 Introduction

Over the past few years, there has been a growing interest in the problem of planning under uncertainty. *Markov decision processes (MDPs)* have received much attention as a basic semantics for this problem. An MDP represents the domain via a set of states, with actions inducing stochastic transitions from one state to another. The key problem with this type of representation is that, in virtually any real-life domain, the state space is quite large. However, many large MDPs have significant internal structure, and can be modeled very compactly if that structure is exploited by the representation. In *factored MDPs*, a state is described implicitly as an assignment of values to some set of *state variables*. A *dynamic Bayesian network (DBN)* [7] can then allow a compact representation of the transition model, by exploiting the fact that the transition of a variable often depends only on a small number of other variables. The momentary rewards can often also be decomposed as a sum of rewards related to individual variables or small clusters of variables.

While these representations allow very large, complex MDPs to be represented compactly, they do not help address the planning problem. Standard algorithms for solving MDPs require the representation and manipulation of *value functions* — functions from the exponentially many states to values. Unfortunately, structure in a factored MDP rarely induces any type of structure in the value function.

An obvious solution is to restrict attention to approximate value functions that can be represented compactly [3]. One very useful approach is to use *linear value functions* — functions that are weighted linear combinations of some small number of *basis functions*. In recent work, there has been some success in using this approach to address the *policy evaluation* problem — determining the value function for a fixed policy. Generally, sampling is used to avoid explicit manipulation of the entire state space [5, 10]. In [9] (KP hereafter), we presented an approach based on approximate dynamic programming. The key to our approach was the use of *factored linear value functions*, where each basis function is restricted to some small subset of the domain variables. We showed that, for a factored MDP and factored value functions, the approximate dynamic programming steps can be implemented in closed form without enumerating the entire state space.

All of these methods compute a linear value function that minimizes error in a weighted least squares sense,

where the (squared) approximation error is weighted by the stationary distribution of the Markov chain induced by the current policy. This means that frequently visited states will have high priority, while infrequently visited states will have only a slight influence on the value function. In a pure prediction context, this is a very natural approximation. However, it is very poorly suited to the task of policy improvement, as the weights can result in very misleading estimates of value in states that are outside the range of the current policy, leading to poor choices in the policy improvement phase.

Our first key result in this paper is a new approach for computing linear value functions that removes the dependence of the error metric on the stationary distribution. In Section 4, we present a closed form set of linear equations whose solution minimizes the Bellman error relative to any set of weights[1]. By divorcing the value determination algorithm from the stationary distribution of the current policy, we can pick an error metric that is more conducive to policy search. Thus, we finally have the capability of using linear value functions for policy iteration.

For the case of factored value functions and a factored MDP, the techniques of KP can be used to generate the equations efficiently, thereby providing an efficient implementation of the value determination step. To construct a full policy iteration algorithm, we must also deal with the issue of representing and manipulating policies over very large state spaces. In Section 7 we show that, for factored value functions and factored MDPs, we can represent the one-step greedy policy compactly as a decision list, and compute its value effectively. The computational cost depends on natural structural parameters of the MDP and the value functions.

When approximately solving an MDP, it is important to evaluate how far our proposed solution is from the optimal. There are known results that allow us to bound this error, but they depend on a *max-norm* bound on the Bellman residual. In Section 8, we present an algorithm that exploits the problem structure to compute max-norm bounds on the Bellman error of a factored value function. This algorithm can be used to bound the overall max-norm error of our approximate value function, and thereby provide guidance on how to adjust our approximation to provide better results.

## 2 Markov Decision Processes

A *Markov Decision Process (MDP)* is defined as a 4-tuple $(S, A, R, P)$ where: $S$ is a set of $N$ states; $A$ is a set of actions; $R$ is a *reward function* $R : S \mapsto \mathbb{R}$, such that $R(s)$ represents the reward obtained by the agent in state $s$; and $P$ is a *transition model* where $P_a(s' \mid s)$ represents the probability of going from state $s$ to state $s'$ with action $a$.

A policy $\pi$ for an MDP is a mapping from $S$ to $A$. It is associated with a *value function* $V^\pi : S \mapsto \mathbb{R}$, where $V^\pi(s)$ is the total cumulative value that the agent gets if it starts at state $s$. We will be assuming that the MDP has an infinite horizon and that future rewards are discounted exponentially with a discount factor $\gamma$. Thus, $V^\pi$ is defined using the fixed point equation:

$$V^\pi(s) = R(s) + \gamma \sum_{s'} P_\pi(s' \mid s) V^\pi(s').$$

It is useful to view this computation from the perspective of matrices and vectors. If we view $V^\pi$ and $R$ as $N$-vectors, and $P_\pi$ as an $N \times N$ matrix, we have the equation

$$V^\pi = R + \gamma P_\pi V^\pi. \tag{1}$$

This is a system of linear equations with one equation for each state, and can be solved easily for small $N$.

There are several ways to find the optimal policy $\pi^*$. A commonly used method is *policy iteration* which repeats the following steps until convergence:

- For our current policy $\pi$, compute $V^\pi$.
- For each action $a$, compute the function $Q_a$:

$$Q_a = R + \gamma P_a V^\pi \tag{2}$$

- Redefine $\pi(s) := \mathrm{argmax}_a Q_a(s)$.

The new policy is called "greedy" with respect to the previous policy and value function because it looks a single step into the future through the $Q_a$ functions. In practice, this process often converges in a very small number of iterations, making it the preferred method for solving MDPs if $V^\pi$ can be computed efficiently.

## 3 Linear Value Functions

In many domains, our state space is very large, and we want to approximate our value functions with more compact ones that can be maintained more easily. A very popular choice is to approximate a value function using *linear regression*. Here, we define our space of allowable value functions $\mathcal{V} \subseteq \mathbb{R}^S$ via a set of *basis functions* $H = \{h_1, \ldots, h_k\}$. A *linear value function* over $H$ is a function $V$ that can be written as $V = \sum_{j=1}^{k} w_j h_j$ for some coefficients $\mathbf{w} = (w_1, \ldots, w_k)$. It is often useful to define an $N \times k$ matrix $A$ whose columns are the $k$ basis functions, viewed as vectors. Our approximate value function is then $A\mathbf{w}$.

---

[1] We note that there are two interpretations of the least squares solution to the Bellman equations. The first is as the direct minimization of the mean-squared Bellman residual error as in [1], while the second is as the fixed point of a Bellman iteration with a least-squares projection of the value function, i.e., the standard linear temporal difference approximation method. We adopt the latter approach in this paper, although our methods can be used for direct minimization of the Bellman residual error as well.

For a given value function $V$, we are often interested in finding the value function $\hat{V} = A\mathbf{w}$ that most closely approximates $V$. The notion of distance that is computationally most convenient is *weighted $L_2$ norm*, where we try to minimize $\sum_s \rho(s)(V(s) - \hat{V}(s))^2$ for some set of non-negative weights $\rho$ that sum to 1 (weighted least squares). We can find the optimal $\hat{V}$ using a simple projection process. Roughly speaking, we define each weight $w_i$ by taking the weighted dot product with the corresponding basis function, and then correcting for the fact that our basis is not orthonormal. More precisely, the projection operation consists of computing $\mathbf{w} = (A^T \Lambda A)^{-1} A^T \Lambda V$, where $\Lambda$ is a weight matrix with diagonal entries equal to our projection weights $\rho$. This operation computes the least-squares projection of $V$ onto the linear space defined by $H$. This computation can be implemented via the weighted dot product operation $(f \bullet g)_\rho$, defined as $\sum_s \rho(s) f(s) g(s)$: The entries of our correction matrix $A^T \Lambda A$ are simply $(h_i \bullet h_j)_\rho$, and the entries of the vector $A^T \Lambda V$ are $(h_i \bullet V)_\rho$. Thus, an efficient implementation of the weighted dot product is the key to the feasibility of this computation.

Now, consider the task of evaluating some policy $\pi$. In this case, our goal is to approximate the true value function $V^\pi$. Unfortunately, we typically do not have $V^\pi$; hence, we usually try to find $\hat{V}$ that minimizes the *Bellman error*: $\hat{V} - (\gamma P_\pi \hat{V} + R)$. In KP, we provided an iterative algorithm for approximate value determination; this approach was aimed at factored MDPs, but applies to the general setting. Let $P_\pi$ be the transition model defined by the policy $\pi$. The iterative value determination equation is

$$V^{(t+1)} = \gamma P_\pi V^{(t)} + R \qquad (3)$$

A used a weighted least-squares approximation to Eq. (3) is:

$$\mathbf{w}^{(t+1)} = (A^T \Lambda A)^{-1} [\gamma A^T \Lambda P_\pi A \mathbf{w}^{(t)} + A^T \Lambda R]$$

Under certain assumptions, this process converges to a fixed point which a bounded distance from the weighted projection of the true value function. One of the key assumptions is that the projection weights $\rho$ must be very close to the stationary distribution of $P_\pi$ (in relative error). This assumption was necessary to ensure a contraction of the iterative algorithm. A similar assumption is also crucial to the other (sampling-based) approaches to the problem [5, 9, 10]. In general, the stationary-weights least-squares approxima-tion has been the only type of approximation error for which theoretical convergence results could be shown.

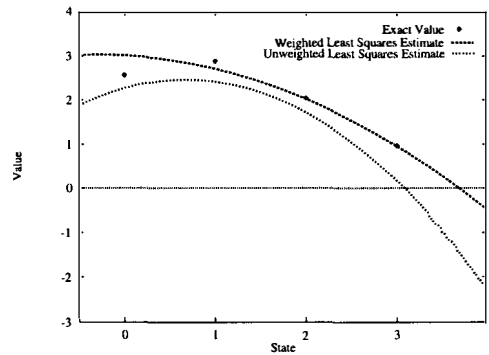

Figure 1: Least squares estimates of the value of policy RRRR.

This approximation is natural in a purely predictive context since it emphasizes the most frequently visited states. However, value determination is often primarily a stepping stone to our ultimate goal, which is the construction of a good policy. Unfortunately, although weighted least-squares is a suitable approximation for predicting the performance of a given policy, it can be extremely unreliable when used as the value determination phase of a policy iteration algorithm.

To understand this issue, consider an MDP with the four states $\{s_0, \ldots, s_3\}$ and the two actions $L$ and $R$. The $R$ action moves "right" — from $s_i$ to $s_{i+1}$ (if available) — with probability 0.9; with probability 0.1 it fails and moves left. The $L$ action has the opposite effect. The two middle states $s_1$ and $s_2$ have reward +1. We specify policies as a string of letters $\pi(s_0)\pi(s_1)\pi(s_2)\pi(s_3)$. The optimal policy for this problem is RRLL.

Suppose that we try to perform policy iteration using weighted least squares approximate value functions, with the basis functions: $h_1(s_x) = 1$, $h_2(s_x) = x$, and $h_3(s_x) = x^2$. If we view value functions as continuous functions over the reals (with $s_x$ representing $x$), our approximate value functions span the space of parabolas.

Assume we start with the policy RRRR, and compute the approximate value function that minimizes the Bellman error relative to the stationary distribution of this policy. The value function is shown in Figure 1. At first, it looks like a reasonable approximation, but it has some critical flaws: The approximation is much better for states 2 and 3 than for states 0 and 1. The reason is that states 0 and 1 are visited very infrequently: for policy RRRR, the stationary distribution is $\rho = [0.00113, 0.01096, 0.09913, 0.88879]$, giving states 0 and 1 very little significance in the weighted least squares fit. The more serious problem is *quali-* *tative*. The shape of the value function is lost, giving state 0 a higher value than state 1. The greedy policy for this value function is LLLL. Symmetrically, the greedy policy for LLLL is RRRR and, thus, policy iteration oscillates between these two suboptimal policies. This phenomenon is not specific to this problem; it has been observed by several researchers on a variety of problems (see, e.g., [2]).

In general, we would prefer to have a $\hat{V}$ that minimizes maximum norm error. Since it is difficult to construct such approximations, a uniform-weighted (unweighted) projection often serves as a tractable substitute. Indeed, Figure 1 also shows the the results of a uniform-weighted value function projection for the policy RRRR. The value function correctly assigns state 0 a lower value than state 1. This leads to a greedy policy of RLLL and the optimal policy of RRLL is found in the following iteration.

## 4 Value determination

As discussed above, all of the approximate value determination procedures proposed so far have relied on the use of stationary-weights least-squares projection to guarantee convergence to a fixed point. In particular, the iterative approximate DP process we used in KP relies on this assumption. In this section, we provide a new approach for computing an approximate linear value function for a given policy $\pi$. The key insight is that an iterative process is not required; we can find the fixed point directly by writing an approximate version of Eq. (1) and solving it:

$$\begin{aligned} A\mathbf{w} &\approx \gamma P_\pi A\mathbf{w} + R \\ A^T \Lambda A \mathbf{w} &\approx A^T \Lambda (\gamma P_\pi A\mathbf{w} + R) \\ \mathbf{w} &\approx (A^T \Lambda A)^{-1} A^T \Lambda (\gamma P_\pi A\mathbf{w} + R) \quad (4) \end{aligned}$$

Letting $B = \gamma (A^T \Lambda A)^{-1} A^T \Lambda P_\pi A$, Eq. (4) is equivalent to $(I - B)\mathbf{w} = (A^T \Lambda A)^{-1} A^T \Lambda R$. As $B$ is a $k \times k$ matrix, we can solve this equation easily if $I - B$ is invertible. Surprisingly, this is almost always the case:

**Theorem 4.1:** *For $B = \gamma (A^T \Lambda A)^{-1} A^T \Lambda P_\pi A$ and $\gamma < 1$, $I - B$ is invertible for all but finitely many $\gamma$.*

**Proof:** The determinant of $I - B$ is a polynomial function in $\gamma$; therefore, it is either uniformly 0 for all $\gamma$ or has only finitely many roots. $I - B$ is invertible for $\gamma = 0$. Hence, the determinant of $I - B$ is not uniformly 0. Therefore, $B$ can fail to be invertible for at most finitely many $\gamma$. ∎

**Corollary 4.2:** *Eq. (4) has a unique solution, which can be computed in closed form.*

Thus, systems without solutions are extremely rare and that if the solution does not exist, a perturbation in $\gamma$ will make the system solvable. Of course, some care must be taken to avoid numerical instability and large errors when inverting matrices that are slightly perturbed from singular matrices.

It is important to point out that the existence conditions for our direct solution are much weaker than the convergence conditions for iterative approximate dynamic programming methods; an iterative solution to Eq. (4) may diverge for almost all starting points even though a unique fixed point solution exists.

The major advantage of our closed form solution is that it can be used to find a weighted least-squares approximation for any weighting $\rho$. Thus, we are free to choose our projection weights to minimize the problem described in Section 3, where using weights corresponding to the stationary distribution of the current policy always misleads us about the value of rarely-visited states. In other words, by allowing different projection weights, we can more evenly distribute our function approximation error and largely overcome the main obstacle to using factored value functions for policy iteration.

This closed form solution defines a computation each of whose operations — the dot product steps — seem to grow linearly with the number of states in the system. Hence, it might appear that this procedure is not particularly helpful: If our state space is small enough to make this computation feasible, it is also small enough to allow an exact solution to the MDP. In the next section, we show that for factored MDPs and factored linear value functions, the relevant dot product operations can be executed exactly in closed form, without an exhaustive enumeration of the (exponentially large) state space.

## 5 Factored MDPs

In a factored MDP, the set of states is described via a set of random variables $\mathbf{X} = \{X_1, \ldots, X_n\}$, where each $X_i$ takes on values in some finite domain $\text{Dom}(X_i)$. A state $\mathbf{x}$ defines a value $x_i \in \text{Dom}(X_i)$ for each variable $X_i$. Thus, the set of states $S = \text{Dom}(\mathbf{X})$ is exponentially large, making it impractical to represent the transition model explicitly as matrices. Fortunately, the framework of *dynamic Bayesian networks (DBNs)* gives us the tools to describe the transition model and reward function concisely.

A Markovian transition model $\tau$ defines a probability distribution over the next state given the current state. Let $X_i$ denote the variable $X_i$ at the current time and $X'_i$ the variable at the next step. The *transition graph* of a DBN is a two-layer directed acyclic graph $G_\tau$ whose nodes are $\{X_1, \ldots, X_n, X'_1, \ldots, X'_n\}$. For simplicity of exposition in the rest of the paper, we assume that $Parents_\tau(X'_i) \subseteq \mathbf{X}$; in graphical terms, all arcs in the DBN are between variables in consecutive time slices. This assumption can be relaxed, but our algorithm becomes somewhat more complex. We denote the parents of $X_i'$ in the graph by $Parents_\tau(X_i')$. Each node $X_i'$ is associated with a *conditional probability distribution (CPD)* $P_\tau(X_i' \mid Parents_\tau(X_i'))$. The transition probability $P_\tau(\mathbf{x}' \mid \mathbf{x})$ is then defined to be $\prod_i P_\tau(x_i' \mid \mathbf{u}_i)$, where $\mathbf{u}_i$ is the value in $\mathbf{x}$ of the variables in $Parents_\tau(X_i')$. Figure 2 shows a DBN with 5 binary state variables. We will use this extremely simple DBN throughout the paper to illustrate some of the concepts we introduce.

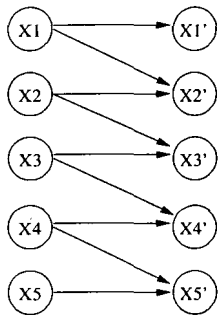

Figure 2: A simple DBN with 5 state variables.

We can define the transition dynamics of an MDP by defining a separate DBN model $\tau_a = \langle G_a, P_a \rangle$ for each action $a$. However, in many cases, different actions have very similar transition dynamics, only differing in their effect on some small set of variables. In particular, in many cases a variable has a default evolution model, which only changes if an action affects it directly [4]. We therefore use the notion of a *default transition model* $\tau_d = \langle G_d, P_d \rangle$. For each action $a$, we define $Effects[a] \subseteq \mathbf{X}'$ to be the variables in the next state whose local probability model is different from $\tau_d$, i.e., those variables $X_i'$ such that $P_a(X_i' \mid Parents_a(X_i')) \neq P_d(X_i' \mid Parents_d(X_i'))$. (We note that $d$ can be an action in our model, in which case $Effects[d] = \emptyset$.) In our example DBN, we will define 5 actions, $a_1 \ldots a_5$ and a default action, $d$. Action $a_i$ changes the CPD of variable $X_i'$, so $Effects[a_i] = \{X_i'\}$.

Finally, we need to provide a compact representation of the reward function. We assume that the reward function is factored additively into a set of localized reward functions, each of which only depends on a small set of variables. For this, and for other reasons, the following definition turns out to be crucial:

**Definition 5.1:** We say that a function $f$ is *restricted* to a domain $\mathbf{C} \subseteq \mathbf{X}$ if $f : Dom(\mathbf{C}) \mapsto I\!R$. If $f$ is restricted to $\mathbf{Y}$ and $\mathbf{Y} \subset \mathbf{Z}$, we will use $f(\mathbf{z})$ as shorthand for $f(\mathbf{y})$ where $\mathbf{y}$ is the part of the instantiation $\mathbf{z}$ that corresponds to variables in $\mathbf{Y}$. ∎

Let $R_1, \ldots, R_r$ be a set of functions, where each $R_i$ is restricted to a cluster of variables $\mathbf{W}_i \subset \{X_1, \ldots, X_n\}$. The reward function associated with the state $\mathbf{x}$ is then defined to be $\sum_{i=1}^{r} R_i(\mathbf{x}) \in I\!R$. For simplicity of notation, we assume that there is a single reward function $R$ that has bounded domain $\mathbf{W}$. (Since our methods are linear, this assumption is totally innocuous.)

One might be led to believe that factored transition dynamics and rewards would result in a structured value function. Unfortunately, this usually is not the case, as shown by KP in Example 2.1. In general, the value function will eventually depend, in an unstructured manner, on all of the variables that have any influence whatsoever, direct or indirect, on a reward.

## 6 Factored value functions

In KP, we observed that, although value functions are not structured, there are many domains where they are "close" to structured. Hence, we might be able to approximate value functions well as a linear combination of functions each of which refers only to a small number of variables. More precisely, we define a value function to be a *factored (linear) value function* if it is a linear value function over the basis $h_1, \ldots, h_k$, where each $h_i$ is restricted to some subset of variables $\mathbf{C}_i$ (as in Definition 5.1). In our simple example DBN, we might have 5 basis functions, $h_1 \ldots h_5$, restricted to $X_1, \ldots, X_5$ respectively. The function $h_i$ would evaluate to 1 if $X_i$ is true and 0 otherwise.

Factored value functions provide the key to doing efficient computations over the exponential-size state sets that we have in factored MDPs. The key insight is that restricted domain functions (including our basis functions) allow certain basic operations to be implemented very efficiently. We now describe these computational building blocks, that are central to our later development.

The first operation is the dot product. Assume that $f$ is restricted to $\mathbf{Y}$ and $g$ is restricted to $\mathbf{Z}$, and let $\mathbf{W} = \mathbf{Y} \cup \mathbf{Z}$. It can easily be shown that

$$(f \bullet g) = \frac{|Dom(\mathbf{X})|}{|Dom(\mathbf{W})|} \sum_{\mathbf{w}} f(\mathbf{w}) \cdot g(\mathbf{w})$$

This computation can be done in time which is linear in $|Dom(\mathbf{W})|$. Assuming that $\mathbf{W}$ is substantially smaller than $\mathbf{X}$, this cost is exponentially lower than the straightforward exhaustive enumeration. In our example, $\mathbf{W} = \{X_1, X_2\}$ for $(h_1 \cdot h_2)$, so there are 4 terms in the summation.

Next, consider a factored transition model $P_\tau$ defined via a DBN $\langle G_\tau, P_\tau \rangle$. A key operation is to *back-project* a function $f$ through $P_\tau$. In linear algebra notation, we want to compute $P_\tau f$, where we view $P_\tau$ as an $N \times N$ matrix (for $N = |S|$) and $f$ as an $N$- vector. The result is a function over $S$. Assume that $f$ is restricted to $\mathbf{Y}$. We define the *back-projection of* $\mathbf{Y}$ *through* $\tau$ as the set of parents of $\mathbf{Y}'$ in the transition graph $G_\tau$; more precisely:

$$\Gamma_\tau(\mathbf{Y}') = \cup_{Y' \in \mathbf{Y}'} Parents_\tau(Y')$$

Now, we can compute

$$\begin{aligned}(P_\tau f)(\mathbf{x}) &= \sum_{\mathbf{x}'} P_\tau(\mathbf{x}' \mid \mathbf{x}) f(\mathbf{x}') \\ &= \sum_{\mathbf{x}'} P_\tau(\mathbf{x}' \mid \mathbf{x}) f(\mathbf{y}') \\ &= \sum_{\mathbf{y}'} P_\tau(\mathbf{y}' \mid \mathbf{x}) f(\mathbf{y}') \sum_{\mathbf{u}' \in \mathbf{x}' - \mathbf{y}'} P_\tau(\mathbf{u}' \mid \mathbf{x}) \\ &= \sum_{\mathbf{y}'} P_\tau(\mathbf{y}' \mid \mathbf{z}) f(\mathbf{y}')\end{aligned}$$

where $\mathbf{z}$ is the value of $\Gamma_\tau(\mathbf{Y})$ in $\mathbf{x}$. Thus, we see that $(P_\tau f)$ is a function whose domain is restricted to $\Gamma_\tau(\mathbf{Y})$. Note that the cost of the computation depends linearly on $|\text{Dom}(\Gamma_\tau(\mathbf{Y}))|$, which in turns depends both on $\mathbf{Y}$, the domain of $f$, and on the complexity of the process dynamics. In our example DBN, the domain of $P_\tau h_2$ is $Parents_\tau(X_1') = \{X_1, X_2\}$.

We can extend this idea to compute the weighted version of these operations, assuming our weights are represented in a factored way. Assume that $\mathbf{X}$ is partitioned into a set of *disjoint* clusters $\mathbf{E}_1, \ldots, \mathbf{E}_q$, such that the weights $\rho$ can be represented as a product of factors (marginals) $\rho_1, \ldots, \rho_k$, where each $\rho_i$ is a factor (distribution) over $\mathbf{E}_i$. It is easily verified that $(f \bullet \rho) = \sum_{\mathbf{y}} f(\mathbf{y}) \rho(\mathbf{y})$. We can easily compute $\rho(\mathbf{y})$ as follows: let $\mathbf{y}_i$ denote the part of $\mathbf{y}$ that overlaps with variables in the cluster $\mathbf{E}_i$. We compute $\rho_i(\mathbf{y}_i)$ by a simple marginalization process, and then multiply the results for all $i = 1, \ldots, q$. We can now use this subroutine for doing the weighted projection, simply by noting that $(f \bullet g)_\rho = ((f \bullet g) \bullet \rho)$.

## 7 Policy Iteration for Factored MDPs

In this section, we show how to implement policy iteration using factored value functions. To make this process tractable, we need to address two fundamental issues. We need to represent policies over an exponentially large space, and we need to show how to perform the computation required for our closed form value determination efficiently. As we now show, we can address both of these issues within the context of our framework.

### 7.1 Policy representation

Assume that we have computed a factored value function $V$ over our basis $H$. Initially, this computation will not be a problem: we can start from some default policy that takes a fixed action $a_0$ in every state, and solve Eq. (4) for $P_{a_0}$, as described in the previous section. If $\mathbf{w}$ is the result, the $V = A\mathbf{w}$ is our factored value function.

Based on $V$, we can easily compute $Q_a$ as in Eq. (2): $Q_a = \gamma P_a A\mathbf{w} + R$. As discussed in Section 6, $P_a A$ can be computed efficiently; it consists of a set of $k$ functions with restricted domains $\Gamma_a(\mathbf{C}_i)$. Thus, $P_a A\mathbf{w}$ is a weighted combination of restricted domain functions. We can compute this $Q_a$ function for every action, including the default transition model $d$.

We now have a set of linear $Q$-functions which implicitly describes a policy $\pi$. It is not immediately obvious that these $Q$ functions will result in a compactly expressible policy. The key insight is that most of the components in the weighted combination will be identical in $P_a A$ and in $P_d A$. Intuitively, a component corresponding to basis function $h_i$ will only be different if the action $a$ influences one of the variables in $\mathbf{C}_i$. More formally, recall that

$$[P_a h_i](\mathbf{x}) = \sum_{\mathbf{c}_i'} P_a(\mathbf{c}_i' \mid \mathbf{z}) h_i(\mathbf{c}_i')$$

where $\mathbf{z}$ is the value of $\Gamma_a(\mathbf{C}_i)$ in $\mathbf{x}$. Now, assume that $Effects[a] \cap \mathbf{C}_i = \emptyset$. In this case, all of the variables in $\mathbf{C}_i$ have the same transition model in $\tau_a$ and $\tau_d$, so that $P_a(\mathbf{c}_i' \mid \mathbf{z}) = P_d(\mathbf{c}_i' \mid \mathbf{z})$ and $[P_a h_i](\mathbf{x}) = [P_d h_i](\mathbf{x})$. Let $I_a$ be the set of indices $i$ such that $Effects[a] \cap \mathbf{C}_i \neq \emptyset$. These are the indices of those basis functions whose back-projection differs in $P_a$ and $P_d$. In our example DBN, actions and basis functions involve single variables, so $I_{a_i} = i$.

We can now define $\delta_a(s) = Q_a(s) - Q_d(s)$. Our analysis shows that $\delta_a(s)$ is a function whose domain is restricted to $\mathbf{T}_a = \cup_{i \in I_a} \Gamma_a(\mathbf{C}_i)$. In our example DBN, $\mathbf{T}_{a_2} = \{X_1, X_2\}$.

Intuitively, we now have a situation where we have a "baseline" value function $Q_d(s)$ which defines a value for each state $s$. Each action $a$ changes that baseline by adding or subtracting an amount from each state. The key point is that this amount depends only on $\mathbf{T}_a$, so that it is the same for all states in which the variables in $\mathbf{T}_a$ take the same values.

We can now define the optimal policy relative to our $Q$ functions. For each action $a$, define a set of *conditionals* $\langle \mathbf{t}, a, \delta \rangle$, where each $\mathbf{t}$ is some assignment of values to the variables $\mathbf{T}_a$, and $\delta$ is $\delta_a(\mathbf{t})$. Now, sort all of the conditionals for all of the actions by order of decreasing $\delta$:

$$\langle \mathbf{t}_1, a_1, \delta_1 \rangle, \langle \mathbf{t}_2, a_2, \delta_2 \rangle, \ldots, \langle \mathbf{t}_L, a_L, \delta_L \rangle \quad (5)$$

Consider our optimal action in a state $\mathbf{x}$. We would like to get the largest possible "bonus" over the default value. If $\mathbf{x}$ is consistent with $\mathbf{t}_1$, we should clearly take action $a_1$, as it will give us bonus $\delta_1$. If not, then we should try to get $\delta_2$; i.e., we should check if $\mathbf{x}$ is consistent with $\mathbf{t}_2$, and if so, take $a_2$. In general, we can view Eq. (5) as a *decision list* representation of a policy, where the optimal action to take in state $\mathbf{x}$ is the action $a_j$ corresponding to the first event $\mathbf{t}_j$ in the list with which $\mathbf{x}$ is consistent.

**Theorem 7.1**: *The optimal one-step lookahead policy for a factored value function $V$ has the form of a decision list as in Eq. (5).*

Note that the number of conditionals in the list is $\sum_a |\text{Dom}(\mathbf{T}_a)|$; $\mathbf{T}_a$, in turn, depends on the set of basis function clusters that intersect with the effects of $a$. Thus, the size of the policy depends in a natural way on the interaction between the structure of our process description and the structure of our basis functions. In our example DBN, the number of conditionals is 18: 2 from $a_1$ and 4 each from $a_2 \ldots a_5$.

### 7.2 Value determination for decision-list policies

The second issue that we must resolve in order to "close the loop" is the computation of the value function for a given policy $\pi$.

The computational ideas described in the previous section allow us to provide an efficient implementation for the value determination algorithm described in Section 4, assuming both the process dynamics and the value function basis are factored. To understand why, consider a factored process $P_\tau$ and a matrix $A$ representing our basis. Recall that the main operation in the value determination process is computing the $k \times k$ matrices $A^T \Lambda A$ and $A^T \Lambda P_\tau A$. We can compute the former by performing $k^2$ weighted dot product operations $(h_i \bullet h_j)_\Lambda$; we can compute the latter by performing the $k^2$ dot product operations $(h_i \bullet P_\tau h_j)_\Lambda$.

The cost of these dot product operations depends on the overlap of the domains of the two functions. (For simplicity of notation, we ignore the weights in our analysis from here on.) More precisely, we define the *structural cost* $cost(A, P_\tau)$ to be

$$\max[\max_{i,j} |\text{Dom}(\mathbf{C}_i \cup \mathbf{C}_j)|, \max_{i,j} |\text{Dom}(\Gamma_\tau(\mathbf{C}_i) \cup \mathbf{C}_j)|].$$

This expression measures the worst-case cost of computing the dot-product of one of our basis functions with another, or with the back-projection of another. It depends on the extent to which the domains of basis functions overlap, and the extent to which back-projection causes them to entangle. (Note that if $X_i$ is always a parent of $X_i'$, the second term in the max is no smaller than the first.) In our example DBN, the structural cost is 8 since $\text{Dom}(\Gamma_\tau(\mathbf{C}_i) \cup \mathbf{C}_j)$ can contain at most 3 binary variables.

We want to apply this idea to the optimal one-step lookahead policy $\pi$ defined in Eq. (5). In other words, we want to solve our fixed point Eq. (4) for a transition model $P_\pi$. In order to apply our efficient dot product operations, the transition model $P_\pi$ must be factored as a DBN. Unfortunately, even with a decision list policy, $P_\pi$ does not have the appropriate structure. Specifically, $P_\pi h_i$ may not have a restricted domain. The solution to this problem is based on the observation that $P_\pi A$ is a combination of $P_{a_l} A$ for the different conditionals in the list $\langle \mathbf{t}_l, a_l, \delta_l \rangle$, with the proportions of the different conditionals being the number of states in which we apply this conditional. As we now show, there is enough structure in $P_\pi h_i$ that we can directly compute entries of $A^T P_\pi A$ efficiently.

We do this by computing $(h_i)^T P_\pi (h_j)$ for each pair of basis functions $h_i, h_j$:

$$(h_i)^T P_\pi h_j = \sum_\mathbf{x} h_i(\mathbf{x})[P_\pi h_j](\mathbf{x})$$

where $[P_\pi h_j](\mathbf{x})$ is the back-projection of $h_j$ through $P_\pi$, evaluated at $\mathbf{x}$. We can partition the states according to the conditionals that are taken in the decision list. For $l = 1, \ldots, L$, let $S_l$ be the set of states in which the conditional $\langle \mathbf{t}_l, a_l, \delta_l \rangle$ is taken. Thus,

$$(h_i)^T P_\pi h_j = \sum_{l=1}^{L} \sum_{\mathbf{x} \in S_l} h_i(\mathbf{x})[P_{a_l} h_j](\mathbf{x}).$$

Consider one of the terms in $\sum_{\mathbf{x} \in S_l} h_i(\mathbf{x})[P_{a_l} h_j](\mathbf{x})$. Recall that $P_{a_l} h_j$ is a restricted domain function whose domain is $\Gamma_{a_l}(\mathbf{C}_j)$. The basis function $h_i$ is also restricted domain, with domain $\mathbf{C}_i$. We can now define $\mathbf{Z}_{a_l,i,j} = \Gamma_{a_l}(\mathbf{C}_j) \cup \mathbf{C}_i$ and rewrite the summation:

$$(h_i)^T P_\pi h_j = \sum_{l=1}^{L} \sum_{\mathbf{z} \in \mathbf{Z}_{a_l,i,j}} h_i(\mathbf{z})[P_{a_l} h_j](\mathbf{z}) \sum_{\mathbf{z} \in S_l:\ \mathbf{Z}_{a_l,i,j}(\mathbf{x}) = \mathbf{z}} 1$$

Let $f_{a_l,i,j}(\mathbf{z})$ be the function $h_i(\mathbf{z})[P_{a_l} h_j](\mathbf{z})$. This is the product of two restricted domain functions and can be computed easily using our techniques in time at most $cost(A, P_{a_l})$. The innermost summation simply counts the number of states in the current partition that are consistent with $\mathbf{z}$. Define this as $N_{a_l,i,j} = |\{\mathbf{x} \in S_l\ :\ \mathbf{Z}_{a_l,i,j}(\mathbf{x}) = \mathbf{z}\}|$. Putting it together, we get:

$$(h_i)^T P_\pi h_j = \sum_{l=1}^{L} \sum_{\mathbf{z} \in \mathbf{Z}_{a_l,i,j}} N_{a_l,i,j} f_{a_l,i,j}(\mathbf{z}). \quad (6)$$

It remains only to compute $N_{a_l,i,j}$. To understand this task, consider the simpler one of counting the number of states in $S_l$. The states in $S_l$ are the ones where we used the $l$'th conditional for selecting our action.

These are states that are consistent with $t_l$, and which are not consistent with $t_1, \ldots, t_{l-1}$. Each $t_m$ is an assignment of values to some set of variables; $x$ can be inconsistent with $t_m$ by being inconsistent with any one of these values. Thus, inconsistency with the previous conditionals can be expressed logically as the conjunction of a set of disjunctions, i.e., a CNF formula. It is fairly straightforward to show, based on this observation, that computing $N_{a_l,i,j}$ is #P-complete (and therefore NP-hard).

Fortunately, this problem will inherit some of its structure from the DBN describing the transition model, and we can use the tools of Bayes net inference to make this computation more tractable. In effect, we can view each statement — that $x$ does not agree with $t_m$ for $i = 1, \ldots, l-1$, that $x$ does agree with $t_l$, and that $x$ agrees with $z$ — as a constraint on $x$. Our goal is to count the number of satisfying instances to this constraint satisfaction problem. Each of these constraints is local, and depends only on a few variables. The constraint on $t_m$ depends only on the variables in $T_{a_m}$, whereas the constraint on $z$ depends only on the variables in $Z_{a_l,i,j}$. We can compute $N_{a_l,i,j}$ in time which is exponential in the induced width of this constraint graph. In our DBN example, we will have one cluster of size 3, arising from the constraint for $Z_{a_l,i,j}$ which involves a pair of variables for $\Gamma_{a_l}(X'_j)$ and one variable for $X_i$, and a chain structure for the constraints on each of the previous action tests. The maximal clique size in a clique tree for this graph is also 3, so that the induced width of resulting constraint graph is at most 2. In general, if each action affects a single variable and the domain of each basis function is also a single variable, then the constraint graph has exactly the same structure as the original DBN (which has the same structure for all actions in this case).

This analysis shows that we can compute the matrix $A^T P_\pi A$. We have already shown how to compute the other parts of Eq. (4), which do not vary with $\pi$. Our result in Theorem 4.1 now applies, so we can compute the value function that is the least-squares approximation to Eq. (4). This value function is factored, allowing us to compute its one-step lookahead policy, thereby closing the loop. Thus, we have shown that we can do policy iteration over these factored value functions and the policies they induce.

## 8  Error Bounds

So far, we have shown how to compute value functions that minimize the (weighted) mean squared Bellman error and then how to use these value functions to select policies in policy iteration. To compute error bounds using standard methods, we need to compute the max-norm error in our value function, which we then can use to bound the distance from our policy to the optimal policy. When we have reached policy $\pi'$, which is the greedy policy for $\hat{V}^\pi$, we compute the maximum Bellman residual error as:

$$\text{BellmanErr}(\hat{V}^\pi) = \max_a \max_\mathbf{x} [Q_a(\mathbf{x}) - \hat{V}^\pi(\mathbf{x})]$$

where $Q_a$ is the one-step greedy Q-function for $\hat{V}^\pi$ as in Section 7. $\hat{V}^\pi$ and each $Q_a$ are sums of restricted domain functions. Hence each inner maximization is over a linear combination of functions, each of which is restricted to some small subset of variables. This type of optimization problem is a *cost network* [8], and can be solved using standard variable elimination algorithms. The computational cost, as for other related structures, is exponential in the induced width of the graph induced by the hyper-edges consisting of the function domains.

For $\text{BellmanErr}(\hat{V}^\pi) \leq \epsilon$, we get [11]:

$$\|V^{\pi^*} - V^{\pi'}\|_\infty \leq \frac{2\epsilon}{1-\gamma}$$

Thus, the *true value* of following $\pi'$ is bounded by a function of the maximum Bellman error of $\hat{V}^\pi$.

The above computation gives us a method of computing the worst-case policy loss for any policy we produce through policy iteration. In general, policy iteration with approximate value functions can produce a sequence of policies of increasing quality. However, approximate policy iteration differs from the exact case in that it can get trapped, repeatedly oscillating through a family of policies without ever finding the globally optimal policy. The loss of the worst policy in this family can be bounded as a function of the worst-case error in the corresponding value functions. [2]. Thus, it is also useful to compute the maximum Bellman error in our policy evaluation phase:

$$\text{BellmanErr}_\pi(\hat{V}) = \max_\mathbf{x} [\hat{V}(\mathbf{x}) - (\gamma(P_\pi \hat{V})(\mathbf{x}) + R(\mathbf{x}))].$$

This can be computed using a combination of the cost-network method and the policy evaluation method described in Section 7. We omit the details for lack of space.

The maximum Bellman error *during* the policy evaluation phase can be used to catch potentially misleading value functions and help correct them. In addition to yielding the maximum Bellman error, the cost network computation tells us the state at which the Bellman error is maximized. If the Bellman error is large, we may wish to change our basis functions, e.g., by adding a basis function that can capture some important correlation. As an alternative, if we are using factored projection weights, we might simply adjust the weights to give the offending state greater importance in the least-squares approximation. Thus, we could gradually adjust our weights with the aim of minimizing the max-norm error in our policy evaluation phase. These methods provide a means of monitoring the quality of the value function approximation during policy iteration, a guide for adjusting the approximation, if necessary, and a means of evaluating any final policy that is selected in comparison to the optimal policy.

## 9 Discussion and future work

In this paper, we have provided a new policy iteration algorithm for factored MDPs, using a factored linear approximation to the value function. A key component of our algorithm is a closed-form value determination method using weighted least squares with arbitrary weights, rather than the stationary distribution weights This method is justified by a theorem showing that the fixed point solution to the approximate dynamic programming equation exists for almost any discount factor. The second key component of our algorithm is the observation that the basic operations can be done effectively in closed form for factored value functions, despite the fact that they are functions over an exponentially large space. This observation also permits the efficient computation of error bounds which, if desired, can be used to adjust the projection weights and evaluate the quality of the resulting policy.

An important theme that recurs throughout our work is the systematic way in which the algorithm exploits the structure of the model. The structure is utilized in many ways: in the operations used for basic value determination, in the compact representation of our decision-list policies, in the counting argument that allowed us to perform value determination for these decision-list policies, and in the computation of the Bellman error. In all of these cases, we saw the same structural features playing the key role: the clusters defined by the domains of the basis functions, their back-projections (for the dynamic programming step) and their forward projections (for the effects of actions). The complexity of our algorithm is determined by the size of these clusters, and by the extent to which they interact with each other: the joint size of overlapping clusters, and the induced width of the graph defined by these clusters. This is a very natural structural property that incorporates properties of the transition dynamics as well as of our chosen basis functions.

This paper opens up many interesting avenues for future work. In one direction, it is clear that we can extend our idea of doing closed-form computations to other MDP solution algorithms, such as linear programming. In a very different direction, we believe that we can extend this approach to exploit various other types of structure in the model, including structured action spaces, where at each stage several actions are taken in parallel, and the context-sensitivity utilized by [3, 6]. As a more ambitious goal, we would also like to extend it to deal with the much harder problem of planning in Partially Observable MDPs.

### Acknowledgments

We thank Xavier Boyen for an insightful discussion that led to Theorem 4.1 and Carlos Guestrin, Uri Lerner and Simon Tong for useful discussions. This work was supported by ARO grant DAAH04-96-1-0341 under the MURI program "Integrated Approach to Intelligent Systems", by the Terman Foundation and by the Sloan Foundation.